# Depth Estimation Algorithm Based on Transformer-Encoder and Feature Fusion

Linhan Xia *
United International College
Computer Science and Technology
Zhu Hai, China
temp.xialinhan@gmail.com*

Junbang Liu
United International College
Computer Science and Technology
Zhu Hai, China
steven.liujunbang@gmail.com

Tong Wu
United International College
Computer Science and Technology
Zhu Hai, China
theawu1010@gmail.com

*Abstract*—This research presents a novel depth estimation algorithm based on a Transformer-encoder architecture, tailored for the NYU and KITTI Depth Dataset. This research adopts a transformer model, initially renowned for its success in natural language processing, to capture intricate spatial relationships in visual data for depth estimation tasks. A significant innovation of the research is the integration of a composite loss function that combines Structural Similarity Index Measure (SSIM) with Mean Squared Error (MSE). This combined loss function is designed to ensure the structural integrity of the predicted depth maps relative to the original images (via SSIM) while minimizing pixel-wise estimation errors (via MSE). This research approach addresses the challenges of over-smoothing often seen in MSE-based losses and enhances the model's ability to predict depth maps that are not only accurate but also maintain structural coherence with the input images. Through rigorous training and evaluation using the NYU Depth Dataset, the model demonstrates superior performance, marking a significant advancement in single-image depth estimation, particularly in complex indoor and traffic environments.

## I. INTRODUCTION

Depth estimation from single images is a challenging yet vital task in computer vision, with applications spanning from augmented reality to autonomous navigation. Recent advancements in deep learning, particularly with transformer models, have opened new avenues for enhancing the accuracy and reliability of depth estimation. This project leverages the NYU and KITTI Depth Dataset, a comprehensive database of indoor scenes, to train and evaluate a specialized depth estimation algorithm. The focus is on utilizing a Transformer-encoder architecture, known for its efficacy in modeling long-range dependencies in data. This approach is expected to improve the depth inference from complex indoor scenes where traditional convolutional methods might struggle.

## II. RELATIVE WORK

Monocular depth estimation, a fundamental task in computer vision, aims to infer depth information from a single image. This task is critical for numerous applications, including autonomous driving, augmented reality, and 3D reconstruction. Over the years, various approaches have been developed, leveraging advancements in machine learning, especially deep learning. These approaches can be broadly categorized into supervised, unsupervised, and semi-supervised methods, each with unique strengths and challenges. This section provides an overview of some of the significant works in this domain, emphasizing the diversity and evolution of techniques ranging from traditional convolutional neural networks to cutting-edge Transformer-based models. The insights gained from these studies not only demonstrate the progress in the field but also highlight the ongoing challenges and potential directions for future research.

### A. Neural Regression Forest by Roy

Roy [1] introduced the Neural Regression Forest (NRF), a novel structure combining a random forest with a convolutional neural network. This approach utilized scanning windows in images as samples, with CNNs filtering these samples stored in random forest tree nodes to produce depth maps. The NRF also incorporated bilateral filtering to address the common problem of lack of smoothness in monocular depth estimation.

### B. Continuous CRFs by Xu

Xu et al. [2] developed a deep learning model integrating Continuous Conditional Random Fields (CRFs) with CNNs. This model efficiently fused continuous CRFs with common CNN architectures and demonstrated the feasibility of a multi-scale fusion approach for monocular depth estimation. However, a significant challenge was the high computational cost associated with CRFs models.

### C. Fully Connected CRFs by Yuan

Yuan [3] addressed the computational challenges of CRFs models by employing fully connected CRFs. By dividing the input into multiple windows and optimizing FC-CRFs within each window, this approach also introduced a multi-head attention mechanism to enhance the accuracy and efficiency of depth estimation.

### D. Unsupervised Learning Framework by Zhan

Zhan [4] proposed an unsupervised learning framework using stereoscopic video sequences. A novel feature reconstruction loss was applied to improve unsupervised monocular depth estimation's accuracy. This framework showed promising results in leveraging unsupervised methods for depth estimation tasks.

### E. Self-Supervised Monodepth2 by Godard

Godard et al. [5] introduced Monodepth2, a self-supervised monocular depth estimation method. This method brought significant innovations to self-supervised learning, such as minimizing reprojection loss, introducing automatic masking loss for stationary pixels, and employing full-resolution multiscale sampling.

*F. Self-Supervised Learning with Self-Attention by Johnston*

Johnston [6] expanded on the Monodepth2 model by integrating a self-attention mechanism and discrete parallax volume (DDV). This method showed excellent performance on the KITTI dataset, surpassing the original Monodepth2 model and highlighting the efficacy of self-supervised learning in monocular depth estimation.

*G. AdaBins with Transformer Network by Farooq*

Farooq et al. [7] utilized the Transformer network, a tool renowned in NLP, for monocular depth estimation. The AdaBins network structure, comprising a depth encoder, a Transformer-based interval predictor, and a depth decoder, presented a cross-domain approach to monocular depth estimation, marking a significant stride in the field.

## III. METHODOLOGY

This research presents an innovative depth estimation algorithm based on the advanced Transformer architecture. The algorithm accepts two forms of inputs: the original image and its frequency-domain representation transformed by Discrete Fourier Transform (DFT). First, the algorithm generates two matrices from each of these two inputs through a downsampling process. Subsequently, these two matrices are fed into the Transformer network to extract and construct two feature matrices, respectively. Next, a feature fusion technique is used to combine these two feature matrices to form a unified information matrix. Eventually, this information matrix is converted into a detailed depth-of-field map by up-sampling process. This algorithm is designed to not only consider the spatial information of the image, but also make full use of its frequency-domain properties, thus improving the accuracy and efficiency of depth estimation.

*A. Data preprocess*

In the initial phase of this study, a Discrete Fourier Transform (DFT) is applied to the input image to transform it into the frequency domain. The mathematical expression for DFT is given as equation (1).

$$F(u,v) = \sum_{x=0}^{M-1} \sum_{y=0}^{N-1} f(x,y) \cdot e^{-2\pi i \left(\frac{ux}{M} + \frac{vy}{N}\right)} \quad (1)$$

where $f(x,y)$ represents the pixel value at position $(x,y)$ in the original image, and $F(u,v)$ denotes the complex value in the frequency domain, representing the frequency component at $(u,v)$. Here, $M$ and $N$ denote the width and height of the image, respectively. This transformation results in two key inputs: the frequency domain image $F(u,v)$ and the original spatial domain image $f(x,y)$. These images are then used as input data for the two branches of the algorithm.

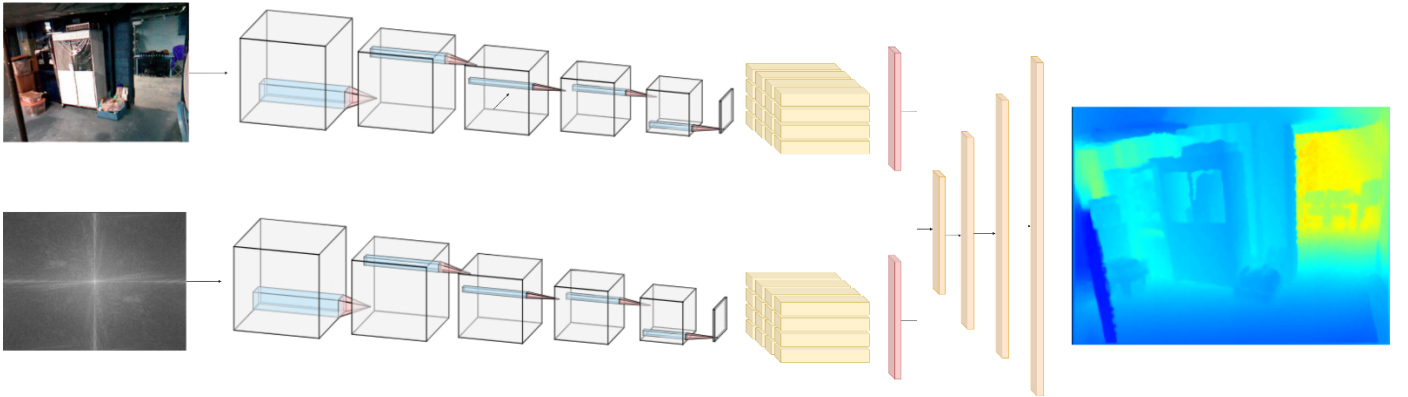

Figure 1. Structure of our work

*B. Encoder Description*

In the proposed depth estimation framework of this study, each image, including both the Discrete Fourier Transform (DFT) processed image and the original image, is individually processed through structurally identical encoders. These encoders are designed based on the Residual Convolutional Neural Network (RCNN) architecture, which is instrumental in extracting feature matrices through a downsampling process.

The key to the encoder's design lies in its utilization of residual connections during the downsampling process. These residual connections aid in preserving significant features of the input data while reducing the dimensions, thereby mitigating the risk of information loss. Formally, if $x$ is the input to a downsampling layer and $F(x)$ represents the transformation by the layer, the output of the layer with a residual connection is given by $x + F(x)$. This approach ensures that the encoder effectively captures intricate details and depth information in the images, providing a rich and accurate feature base for subsequent depth estimation.

Each encoding block in the encoder includes convolutional layers, batch normalization, and the ReLU activation function. The convolutional layers are responsible for feature extraction. The batch normalization is mathematically represented as equation (2.3).

$$\hat{x}^{(k)} = \frac{x^{(k)} - \mu_{\mathcal{B}}}{\sqrt{\sigma_{\mathcal{B}}^2 + \epsilon}} \quad (2)$$

$$y^{(k)} = \gamma^{(k)} \hat{x}^{(k)} + \beta^{(k)} \quad (3)$$

where $x^{(k)}$ is the input to the batch normalization layer, $\mu_B$ and $\sigma_B^2$ are the mean and variance of the inputs, $\gamma^{(k)}$ and $\beta^{(k)}$ are parameters to be learned, and $\epsilon$ is a small constant added for numerical stability. Batch normalization helps in accelerating the training speed and enhancing the model's generalization capability.

The ReLU activation function is defined as equation (4).

$$ReLU = \max(0, x) \quad (4)$$

This function introduces non-linearity to the network, enabling the model to learn more complex feature representations.

*C. Feature matrix processing*

In this study, the feature matrices obtained from the encoders are first subdivided into 16 smaller image patches (patches). This step aims to decompose the larger feature matrices into more manageable units, facilitating the efficient processing by the Transformer architecture. Subsequently, each image patch is processed through the Transformer architecture, which includes both self-attention and multi-head attention mechanisms.

The self-attention mechanism enables the model to consider other elements of the sequence while processing each element of the sequence, with the formula given as equation (5).

$$Attention(Q, K, V) = softmax\left(\frac{QK^T}{\sqrt{d_k}}\right)V \quad (5)$$

where $Q$, $K$, and $V$ respectively represent the query, key, and value, and $d_k$ is the dimension of the key.

The multi-head attention mechanism, on the other hand, disperses the self-attention across multiple heads, allowing each head to independently learn different parts of the input data. The formula for multi-head attention as equation (6,7).

$$MultiHead(Q, K, V) = Concat(head_1, \dots, head_h)W^O \quad (6)$$

$$head_i = Attention(QW_i^Q, KW_i^K, VW_i^V) \quad (7)$$

Here, $W_i^Q$, $W_i^K$, $W_i^V$, and $W^O$ are trainable weight matrices, and $h$ represents the number of heads.

Through this mechanism, the model in this study effectively captures the complex relationships both within and between different image patches. After processing, these Transformer-processed patches are reassembled into a complete feature matrix, laying a solid foundation for subsequent upsampling steps and the reconstruction of the depth image. This design enables the model to delve into deeper features of the image, thereby providing more precise information for depth estimation.

*D. Feature Fusion*

In the depth estimation algorithm of this study, the outputs from the frequency domain image and the original image are combined through a feature fusion process based on an attention mechanism. This step is crucial to the algorithm as it merges information from both the frequency and spatial domains, thereby enhancing the accuracy of depth estimation. The specific process of feature fusion can be described as follows:

Firstly, the similarity between the feature matrix from the frequency domain $F_{freq}$ and the feature matrix from the original image $F_{orig}$ is calculated. Then, the softmax function is applied to these similarities to obtain attention weights. The formula is as equation (8).

$$A = softmax(F_{freq} \cdot F_{orig}^T) \quad (8)$$

Finally, these attention weights $A$ are used to weight the feature matrix $F_{orig}$ from the original image, resulting in the fused feature matrix $F_{fused}$. The fusion process can be represented by the equation (9).

$$F_{fused} = A \cdot F_{orig} \quad (9)$$

This process ensures that important features from both domains are effectively combined, providing a more comprehensive set of feature information for depth estimation.

*E. Decoder Description*

In this research, the depth estimation framework involves up-sampling fused feature data to transform it from low-resolution back to its original high-resolution size, focusing on preserving and enhancing essential features. Techniques like Transpose Convolution or Pixel Shuffle are used to increase the image size while minimizing feature distortion. Once restored to its original size, the image undergoes further post-processing, including filtering and feature enhancement, to improve detail and depth perception while avoiding extra noise. This process ultimately yields a final image that closely resembles the target in appearance and offers superior depth information.

*F. Description of Loss Function*

In this study, we employ a composite loss function combining Mean Squared Error (MSE) and Structural Similarity Index (SSIM) for training the depth estimation model. This design aims to consider both pixel-level accuracy and overall structural consistency of images, achieving more accurate and visually pleasing depth estimation.

The loss function is comprised of two components, with their mathematical expressions as follows.

MSE is defined as equation (10).

$$MSE(pred, target) = \frac{1}{n}\sum_{i=1}^{n}(pred - target)^2 \quad (10)$$

where $pred$ are the predicted values, $target$ are the true values, and $n$ is the number of pixels in the images.

SSIM is defined as equation (11).

$$SSIM = \frac{(2\mu_p\mu_t + C1)(2\sigma_{p,t} + C2)}{(\mu_p^2 + \mu_t^2 + C1)(\sigma_p^2 + \sigma_t^2 + C2)} \quad (11)$$

where $\mu_p$ and $\mu_t$ are the average intensities of the predicted and target images, $\sigma_p$ and $\sigma_t$ are the variances, $\sigma_{p,t}$ is the covariance, and $C1$ and $C2$ are constants to stabilize the division.

The overall loss function is then given as equation (12).

$$Loss = (1 - \alpha) \cdot MSE + \alpha \cdot (1 - SSIM) \quad (12)$$

Here, the hyperparameter \( \alpha \) is used to balance the trade-off between MSE and SSIM, and can be adjusted experimentally for optimal performance.

Through this composite loss function, our model effectively combines pixel-level accuracy with structural image integrity, enhancing the depth estimation's overall effectiveness and visual quality.

## IV. Performance Evaluation

In the previous section, we proposed to combine the SSIM loss function with the MSE loss function, this method can consider both the pixel accuracy and structural similarity of the image, improving the accuracy of our depth estimation. The purpose of this section is to find the best balance point to optimize the performance of the model by adjusting the weight $\alpha$ between SSIM and MSE. In addition, all our tests are performed on the NYU-Depth V2 dataset and KITTI dataset.

### A. Metrics

This research apply four common metrics to evaluate the performance of the model under different parameter settings: Absolute Relative Difference (Abs Rel), Squared Relative Difference (Sq Rel), Root Mean Squared Error (RMSE), and Root Mean Squared Error Log (RMSE log). The smaller the value of these four metrics, the better the model performs on the database.

### B. Implementation details

In this research, we have implemented our proposed network using the PyTorch framework and with the help of NVIDIA A100 GPU computational resources on the Google Colab platform. For the training model process, we chose to use the Adam optimizer with an initial learning rate of $1 \times 10^{-4}$. Because of the large KITTI dataset and NYU dataset we chose and the limitation of computing resources, we set the total number of epochs for the training process to 20 and the batch size to 16. In addition, we used six different $\alpha$ values (0.3, 0.4, 0.5, 0.6, 0.7, 0.8) to test the model. These $\alpha$ values adjusted the share of structural similarity index (SSIM) loss and mean square error (MSE) loss in the total loss. For example, with an $\alpha$ value of 0.3, the SSIM loss accounts for 30% of the total loss, while the MSE loss accounts for 70%, as shown in Table 1.

The results of testing the trained model on the KITTI dataset are shown in Table 2, and the results of testing the model on the NYU dataset are shown in Table 3. By testing on these two datasets, we observe that when the $\alpha$ value is set to 0.8, the model performs well on both datasets, outperforming the other settings. It is also worth noting that with a value of 0.8 for $\alpha$, the values of the four performance evaluation metrics obtained by the model tested on the NYU dataset are better than the results obtained on the KITTI dataset, which suggests that the model performs more accurately for depth estimation in outdoor conditions than in indoor environments. In order to be able to observe the performance of the model more intuitively, we visualized the experimental results as shown in tables.

### C. Comparison with other models

To validate the effectiveness of our model, we compare the main research results in the field of monocular depth estimation in recent years. The performance comparison results on the official NYU-Depth-v2 test set are shown in Table 4. Although the performance of most monocular depth estimation models on the NYU-Depth-v2 test set seems to be saturated in the long run, our model still has a small but significant advantage over the other models in all key metrics. In addition, Table 5 shows the results of our performance comparison on the KITTI dataset. Our proposed architecture significantly outperforms the previous state-of-the-art in all evaluation metrics. This result demonstrates that using a network architecture based on a transformer encoder and feature fusion is not only effective but also opens up new research paths for monocular depth estimation applications.

TABLE I.  EXPERIMENTAL RESULTS ON THE NYU-DEPTH-V2 DATASET

| $\alpha$ value | RMSE | Abs Rel |
|---|---|---|
| 0.3 | 0.827 | 0.217 |
| 0.4 | 0.682 | 0.139 |
| 0.5 | 0.467 | 0.129 |
| 0.6 | 0.403 | 0.121 |
| 0.7 | 0.386 | 0.117 |
| **0.8** | **0.381** | **0.115** |

TABLE II.  EXPERIMENTAL RESULTS ON THE KITTI DATASET

| $\alpha$ value | RMSE | Abs Rel |
|---|---|---|
| 0.3 | 4.950 | 0.119 |
| 0.4 | 4.494 | 0.104 |
| 0.5 | 3.076 | 0.082 |
| 0.6 | 2.972 | 0.077 |
| 0.7 | 2.727 | 0.073 |
| **0.8** | **2.683** | **0.072** |

## V. Conclusions

In this research, we have developed a depth estimation algorithm that leverages the Transformer-encoder architecture, tailored for use with the NYU and KITTI Depth Datasets. Our approach, inspired by the Transformer model's success in natural language processing, focuses on capturing complex spatial relationships in visual data to enhance depth estimation accuracy. A notable feature of our method is the adoption of a composite loss function combining the Structural Similarity Index Measure (SSIM) and Mean Squared Error (MSE). This function aims to balance the accuracy of depth map predictions at both structural and pixel levels.

Our model was rigorously trained and evaluated using the NYU Depth Dataset, showing promising results, particularly in complex indoor environments. An integral part of our research was exploring the optimal balance in the loss function by adjusting the weight $\alpha$ between SSIM and MSE, with an $\alpha$ value of 0.8 showing good performance on both datasets.

Overall, this research contributes to the field of depth estimation by exploring the application of combination of SSIM and MSE loss function in this domain. The results are encouraging and suggest potential for future research, although they should be seen as an initial step rather than a conclusive finding in the broader context of depth estimation technologies.

TABLE III. COMPARISON OF PERFORMANCES ON THE NYU-DEPTH-V2 DATASET

| Method | Encoder | RMSE | Abs Rel | Sq Rel | RMSE log |
|---|---|---|---|---|---|
| DenseDepth [8] | DenseNet-169 | 0.465 | 0.118 | - | 0.171 |
| VNL [9] | MobileNetV2 | 0.485 | 0.134 | - | 0.185 |
| PAP-Depth [10] | ResNet-18 | 0.497 | 0.121 | - | 0.175 |
| SDC-Depth [11] | ResNet-50 | 0.497 | 0.128 | - | 0.174 |
| **Our model** | **ViT-hybrid** | **0.381** | **0.115** | **0.12** | **0.165** |

TABLE IV. VISUALIZING DEPTH ESTIMATION RESULTS ON THE KITTI DATASET

| Method | Encoder | RMSE | Abs Rel | Sq Rel | RMSE log |
|---|---|---|---|---|---|
| DenseDepth [8] | DenseNet-169 | 4.17 | 0.093 | 0.589 | 0.168 |
| VNL [9] | MobileNetV2 | 3.258 | 0.072 | - | 0.124 |
| SC-Depth [12] | ResNet-18 | 4.95 | 0.119 | 0.857 | 0.197 |
| SC-Depth [12] | ResNet-50 | 4.706 | 0.114 | 0.813 | 0.191 |
| **Our model** | **ViT-hybrid** | **2.683** | **0.072** | **0.179** | **0.119** |

REFERENCES

[1] A. Roy and S. Todorovic, "Monocular Depth Estimation Using Neural Regression Forest," Jun. 2016, doi: https://doi.org/10.1109/cvpr.2016.594.
[2] D. Xu, E. Ricci, W. Ouyang, X. Wang, and Nicu Sebe, "Multi-scale Continuous CRFs as Sequential Deep Networks for Monocular Depth Estimation," Jul. 2017, doi: https://doi.org/10.1109/cvpr.2017.25.
[3] W. Yuan, X. Gu, Z. Dai, S. Zhu, and P. Tan, "Neural Window Fully-connected CRFs for Monocular Depth Estimation," Jun. 2022, doi: https://doi.org/10.1109/cvpr52688.2022.00389.
[4] H. Zhan, R. Garg, Chamara Saroj Weerasekera, K. Li, H. Agarwal, and I. R. Reid, "Unsupervised Learning of Monocular Depth Estimation and Visual Odometry with Deep Feature Reconstruction," Computer Vision and Pattern Recognition, Jun. 2018, doi: https://doi.org/10.1109/cvpr.2018.00043.
[5] C. Godard, Oisin Mac Aodha, M. Firman, and G. J. Brostow, "Digging Into Self-Supervised Monocular Depth Estimation," International Conference on Computer Vision, Oct. 2019, doi: https://doi.org/10.1109/iccv.2019.00393.
[6] A. Johnston and G. Carneiro, "Self-Supervised Monocular Trained Depth Estimation Using Self-Attention and Discrete Disparity Volume," Computer Vision and Pattern Recognition, Jun. 2020, doi: https://doi.org/10.1109/cvpr42600.2020.00481.
[7] Shariq Farooq Bhat, Ibraheem Alhashim, and P. Wonka, "AdaBins: Depth Estimation Using Adaptive Bins," arXiv (Cornell University), Jun. 2021, doi: https://doi.org/10.1109/cvpr46437.2021.00400.
[8] L. Lin, G. Huang, Y. Chen, L. Zhang, and B. He, "Efficient and High-Quality Monocular Depth Estimation via Gated Multi-Scale Network," IEEE Access, vol. 8, pp. 7709–7718, 2020, doi: https://doi.org/10.1109/ACCESS.2020.2964733.
[9] Wei Hsian Yin, Y. Liu, C. Shen, and Y. Yan, "Enforcing Geometric Constraints of Virtual Normal for Depth Prediction," Oct. 2019, doi: https://doi.org/10.1109/iccv.2019.00578.
[10] Z. Zhang, Z. Cui, C. Xu, Y. Yan, Nicu Sebe, and J. Yang, "Pattern-Affinitive Propagation Across Depth, Surface Normal and Semantic Segmentation," Institutional Research Information System (Università degli Studi di Trento), Jun. 2019, doi: https://doi.org/10.1109/cvpr.2019.00423.
[11] L. Wang, J. Zhang, O. Wang, Z. Lin, and H. Lu, "SDC-Depth: Semantic Divide-and-Conquer Network for Monocular Depth Estimation," Jun. 2020, doi: https://doi.org/10.1109/cvpr46600.2020.00062.
[12] S. Bhat et al., "Adabins: Depth Estimation Using Adaptive Bins," 2021. [Online]. Available: https://doi.org/10.1109/cvpr46437.2021.00400
[13] Yang, J., An, L., Dixit, A., Koo, J., & Park, S. I. (2022). Depth estimation with simplified transformer. arXiv preprint arXiv:2204.13791.
[14] Karpov, Aleksei, and Ilya Makarov. "Exploring efficiency of vision transformers for self-supervised monocular depth estimation." 2022 IEEE International Symposium on Mixed and Augmented Reality (ISMAR). IEEE, 2022.
[15] Han, D., Shin, J., Kim, N., Hwang, S., & Choi, Y. (2022). Transdssl: Transformer based depth estimation via self-supervised learning. IEEE Robotics and Automation Letters, 7(4), 10969-10976.
[16] Varma, Arnav, et al. "Transformers in self-supervised monocular depth estimation with unknown camera intrinsics." arXiv preprint arXiv:2202.03131 (2022).
[17] Tang, Shuai, et al. "CATNet: Convolutional attention and transformer for monocular depth estimation." Pattern Recognition 145 (2024): 109982.
[18] Agarwal, Ashutosh, and Chetan Arora. "Depthformer: Multiscale vision transformer for monocular depth estimation with global local information fusion." 2022 IEEE International Conference on Image Processing (ICIP). IEEE, 2022.
[19] Tomar, Snehal Singh, Maitreya Suin, and A. N. Rajagopalan. "Hybrid Transformer Based Feature Fusion for Self-Supervised Monocular Depth Estimation." European Conference on Computer Vision. Cham: Springer Nature Switzerland, 2022.
[20] Mertan, Alican, Damien Jade Duff, and Gozde Unal. "Single image depth estimation: An overview." Digital Signal Processing 123 (2022): 103441.
[21] Xu, Haofei, et al. "Unifying flow, stereo and depth estimation." IEEE Transactions on Pattern Analysis and Machine Intelligence (2023).
[22] Li, Zhenyu, et al. "Depthformer: Exploiting long-range correlation and local information for accurate monocular depth estimation." Machine Intelligence Research 20.6 (2023): 837-854.
[23] Masoumian, Armin, et al. "Gcndepth: Self-supervised monocular depth estimation based on graph convolutional network." Neurocomputing 517 (2023): 81-92.
[24] Logothetis, Fotios, et al. "A CNN based approach for the point-light photometric stereo problem." International Journal of Computer Vision 131.1 (2023): 101-120.